\pdfoutput=1

\documentclass[11pt]{article}

\usepackage[final]{acl}

\usepackage{times}
\usepackage{latexsym}
\usepackage{booktabs}
\usepackage{amsmath}
\usepackage{amssymb}
\usepackage{amsfonts}
\usepackage{amsthm}
\usepackage{graphicx}
\usepackage{subfig}
\usepackage{fancyvrb}
\usepackage{listings}
\usepackage[T1]{fontenc}

\usepackage[utf8]{inputenc}

\usepackage{bm}

\usepackage{microtype}

\usepackage{inconsolata}

\usepackage{graphicx}

\usepackage[disable]{todonotes}
\makeatletter
\newcommand*\iftodonotes{\if@todonotes@disabled\expandafter\@secondoftwo\else\expandafter\@firstoftwo\fi}
\makeatother

\newcommand{\note}[4][]{{\todo[author=#2,color=#3,size=\scriptsize,fancyline,caption={},#1]{#4}}}

\newcommand{\response}[1]{\vspace{3pt}\hrule\vspace{3pt}\textbf{#1:}}

\newcommand{\jason}[2][]{\note[#1]{jason}{green!40}{#2}}

\newcommand{\leo}[2][]{\note[#1]{leo}{purple!40}{#2}}

\usepackage{xcolor}
\usepackage{fancyvrb}
\usepackage{adjustbox}
\usepackage{array}
\usepackage{enumitem}
\usepackage[capitalize,noabbrev]{cleveref}

\crefname{section}{\S}{\S\S}
\Crefname{section}{\S}{\S\S}
\crefformat{section}{#2\S#1#3}
\Crefformat{section}{#2\S#1#3}
\crefrangeformat{section}{\S\S#3#1#4--#5#2#6}
\Crefrangeformat{section}{\S\S#3#1#4--#5#2#6}
\crefmultiformat{section}{\S#2#1#3}{ and~\S#2#1#3}{, \S#2#1#3}{ and~\S#2#1#3}
\Crefmultiformat{section}{\S#2#1#3}{ and~\S#2#1#3}{, \S#2#1#3}{ and~\S#2#1#3}
\crefrangemultiformat{section}{\S\S#3#1#4--#5#2#6}{ and~\S\S#3#1#4--#5#2#6}{, \S\S#3#1#4--#5#2#6}{ and~\S\S#3#1#4--#5#2#6}
\Crefrangemultiformat{section}{\S\S#3#1#4--#5#2#6}{ and~\S\S#3#1#4--#5#2#6}{, \S\S#3#1#4--#5#2#6}{ and~\S\S#3#1#4--#5#2#6}
\crefname{table}{Table}{Tables}
\crefname{figure}{Fig.}{Figs.}
\crefname{algorithm}{Alg.}{Algs.}
\crefname{equation}{Eq.}{Eqs.}
\crefname{definition}{Def.}{Definitions}
\crefname{appendix}{App.}{Appendices}
\crefname{theorem}{Thm.}{Theorems}
\crefname{myexample}{Example}{Examples}
\crefname{prop}{Prop.}{Propositions}
\crefname{cor}{Corollary}{Corollaries}
\crefname{observation}{Observation}{Observations}
\crefname{assumption}{Assumption}{Assumptions}
\crefname{hypothesis}{Hyp.}{Hypotheses}

\title{LLMs Know More About Numbers than They Can Say}

\author{Fengting Yuchi$^{1}$\;~Li Du$^{2}$~\;~Jason Eisner$^2$ \\
    $^1$Shanghai Jiao Tong University \and $^2$Johns Hopkins University \\
  \href{mailto:yc1114@sjtu.edu.cn}{\texttt{yc1114@sjtu.edu.cn}} \and
  \{\href{mailto:leodu@cs.jhu.edu}{\texttt{leodu}}\texttt{,}\href{mailto:jason@cs.jhu.edu}{\texttt{jason}}\}\texttt{@cs.jhu.edu}
}

\begin{document}
\maketitle
\begin{abstract}

Although state-of-the-art LLMs can solve math problems, we find that they make errors on numerical comparisons with mixed notation: ``Which is larger, $5.7 \times 10^2$ or $580$?''
This raises a fundamental question: Do LLMs even know how big these numbers are?
We probe the hidden states of several smaller open-source LLMs.  A single linear projection of an appropriate hidden layer encodes the \emph{log-magnitudes} of both kinds of numerals, allowing us to recover the numbers with relative error of about 2.3\% (on restricted synthetic text) or 19.06\% (on scientific papers).
Furthermore, the hidden state after reading a \emph{pair} of numerals encodes their \emph{ranking}, with a linear classifier achieving over 90\% accuracy.
Yet surprisingly, when explicitly asked to rank the same pairs
of numerals, these LLMs achieve only 50--70\% accuracy, with worse performance for models whose probes are less effective.
Finally, we show that incorporating the classifier probe's log-loss as an auxiliary objective during finetuning brings an additional 3.22\% improvement in verbalized accuracy over base models, demonstrating that improving models' internal magnitude representations can enhance their numerical reasoning capabilities.
Our code is available at \href{https://github.com/VCY019/Numeracy-Probing}{\texttt{github.com/VCY019/Numeracy-Probing}}.

\end{abstract}

\section{Introduction}

\begin{table}[t]
\centering
\small
\adjustbox{max width=\columnwidth}{%
\begin{tabular}{lcc}
\toprule
\textbf{Model} &
\textbf{\shortstack{Verbalization \\ Accuracy}} &
\textbf{\shortstack{Probing \\ Accuracy}}
\\
\midrule
GPT-4.1 &
94.19 & ---
\\
GPT-4.1-mini &
92.12 & ---
\\ 
\midrule
DeepSeek-R1-Distill-Llama-8B &
57.81 &
95.62
\\
DeepSeek-R1-Distill-Qwen-7B &
64.19 & 
97.38
\\
Llama-2-7b &
50.81 &
98.44
\\
Llama-3.1-8B-Instruct &
55.06 &
94.81
\\
Mistral-7B-v0.1 &                   
50.00 &
96.44 
\\
OLMo-2-1124-7B &                    53.44 &
93.50
\\
Qwen3-8B &           
70.00 &
98.88
\\
\bottomrule
\end{tabular}%
}
\caption{Cross-notation comparison task on held-out data (see \cref{appendix:synthetic-data-setup} for data details). 
\textbf{Verbalization accuracy} evaluates verbal responses using a one-shot prompt and no training, while \textbf{probing accuracy} measures performance of a linear classifier trained on hidden states. All values are percentages without the \% sign. 
Despite hidden states containing rather accurate comparison information extractable via linear probes, verbalization performs substantially worse.}
\label{tab:verbal vs. probing}
\end{table}%

\begin{figure*}[ht]
    \centering
    $\!$\subfloat[Synth, dec \textcolor{gray}{(18)}]{
     \includegraphics[width=0.15\textwidth]{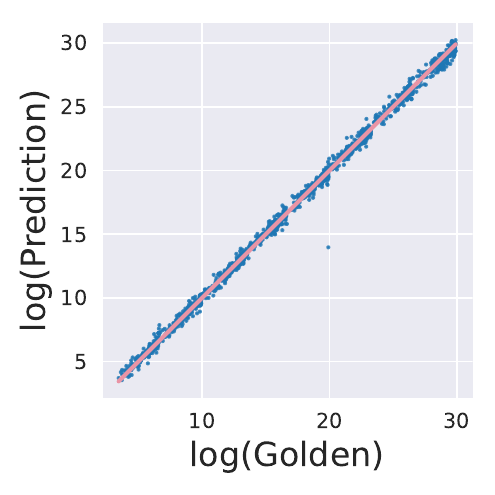}
    \label{fig:scatter-dec-synthetic}
    }
    $\!$\subfloat[Synth, sci \textcolor{gray}{(31)}]{
        \includegraphics[width=0.15\textwidth]{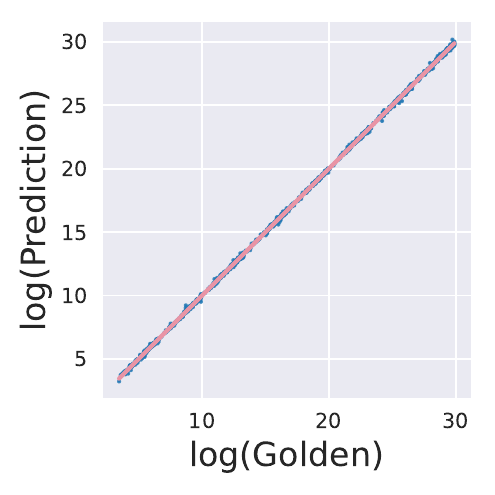}
    \label{fig:scatter-sci-synthetic}
    } 
    $\!$\subfloat[Synth, mixed \textcolor{gray}{(30)}]{
        \includegraphics[width=0.15\textwidth]{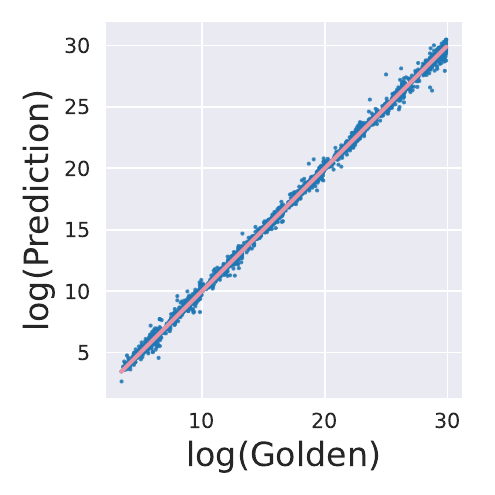}
        \label{fig:scatter-mix-synthetic}
    } 
    $\!$\subfloat[ArXiv, dec \textcolor{gray}{(5)}]{
        \includegraphics[width=0.16\textwidth]{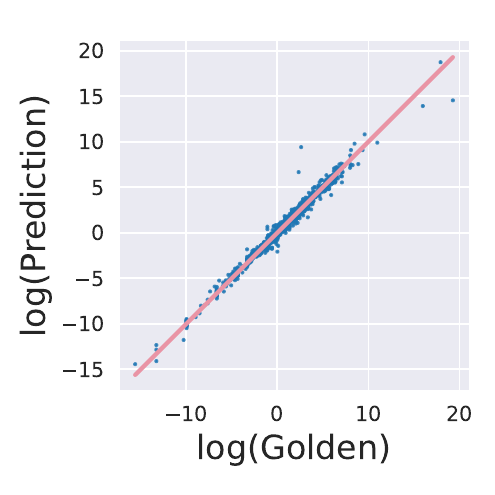}
    }
    $\!$\subfloat[ArXiv, sci \textcolor{gray}{(5)}]{
        \includegraphics[width=0.16\textwidth]{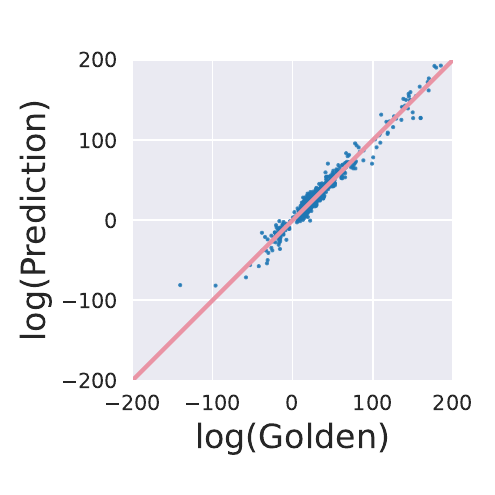}
    }
    $\!$\subfloat[ArXiv, mixed \textcolor{gray}{(10)}]{
        \includegraphics[width=0.16\textwidth]{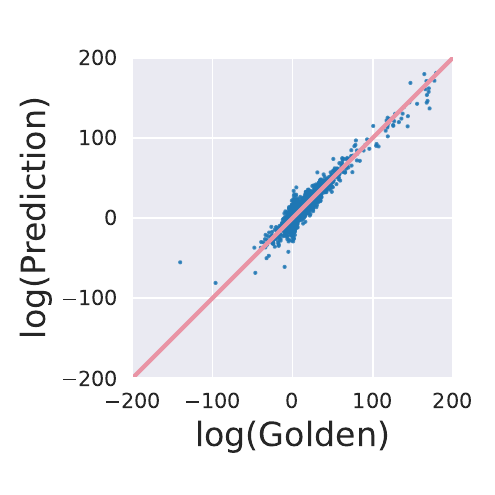}
        \label{fig:scatter-mix-arxiv}
    }
    \caption{
        Scatterplots of predicted vs.\@ true (``golden'') log-magnitudes for Mistral-7B (32 layers) across different datasets and notation types.
        ``Synth'' refers to the synthetic cross-notation data we constructed in \cref{sec:experimental-setup}.
        Dec/sci/mixed conditions train and test on decimals only, scientific notation numerals only, and both, respectively.  
        We train and test a probe at each layer, and then plot performance on held-out test data for only the probe that achieved the highest $R^2$ on held-out validation data.  
        The parenthesized number is the layer index of that probe.
    }
    \label{fig:scatter-all}
\end{figure*}

Large language models (LLMs) are increasingly used in mathematical \citep{jiang2023mistral7b}, scientific \citep{taylor2022galacticalargelanguagemodel,zhang2025scientific}, financial \citep{wu2023bloomberggptlargelanguagemodel}, and engineering \citep{jimenez2024swebench} domains.  In these domains, numeracy—the ability to understand and reason about \emph{numbers}—is an essential basic skill.

While LLMs can correctly answer questions such as ``What is $5.7 \times 10^2$?'' and ``Which is larger, 570 or 580?'', we find that even leading non-reasoning models like GPT-4.1 make many mistakes\jason{changed from "perform disastrously" since GPT-4.1 gets only 7\% error} on cross-notation comparisons like ``Which is larger, $5.7 \times 10^2$ or $580$?''
This discrepancy raises questions: Do LLMs know how big these numbers are, representing them internally in a way that supports intelligent discussion of scientific text?  Or do they lack number sense and are specialized to specific manipulations of \emph{numerals}---the textual representations of numbers---in specific notations?\leo{cite mistral in intro for math problem solving \response{fengting} cited in the first sentence of introduction, not sure whether it's the suitable place\response{jason} We should probably say explicitly whether the specific models we test on comparisons are good at other math tasks.  Maybe this is what Leo was suggesting?}

To study these questions, we first probe the internal numerical knowledge of smaller LLMs (7B--8B parameters) by training linear probes on hidden states of numeral tokens, following \citet{zhu-etal-2025-language}.
We train and test on two kinds of text: synthetic prompts and real-world arXiv papers.
We find that standard linear regression can successfully predict log-magnitudes
(\cref{sec:regression-probing}).  For example, Mistral-7B has layers that linearly encode numerals to \textasciitilde2.3\% median relative error on synthetic data and \textasciitilde19.06\% on scientific papers.

Second, we show that these smaller LLMs also internally \emph{compare} numbers (\cref{sec:classification-probing}).  A linear binary classifier (trained with logistic regression) reveals which of the two numbers in the question is bigger, even when the numbers are so close that their predicted magnitudes (using the previous method) are too noisy for reliable comparison.

Despite having done these computations internally, these LLMs struggle significantly at \emph{verbalizing} the cross-notation comparison (\cref{sec:verbalization-results}).
When prompted to say which of two cross-notation numerals is larger, the models achieve only 50--70\% accuracy—far closer to the 50\% random baseline than 
their internal knowledge would predict.
This sharp performance gap indicates that LLMs possess numerical knowledge that they cannot reliably verbalize through language generation.

Still, probing effectiveness in early layers does \emph{correlate} well with verbalized performance (\cref{sec:correlation}).  For a probe of either type (applied to the first 3 layers), probe accuracy predicts the accuracy of verbal output (produced by the last layer) across diverse 7B--8B LLMs (\cref{fig:correlation-regression-mse,fig:correlation-classification,fig:correlation-regression-all}).

To test whether this connection is causal, we experiment with adding the classifier probe's training loss to the LLM's log-loss training objective on the task of verbalized cross-notation comparison (\cref{sec:finetuning}).
Most LLMs do achieve higher verbalized comparison accuracy when fine-tuned with this augmented objective
rather than the standard objective
(\cref{tab:finetune-accuracy}). 
This finding suggests that targeting internal representations is a promising approach to improving numeracy in LLMs.  That is, models know more than they can say---but improving what they know improves what they can learn to say.

To summarize our key findings:
\begin{itemize}[nosep]
\item Numerical log-magnitudes and magnitude comparisons are encoded linearly, for various LLMs, datasets, and numeral notations.
\item Even so, LLMs verbalize cross-notation comparisons poorly.
\item 
Tuning a model's internal representations to 
better encode numeric comparisons helps the model better answer such comparison questions verbally.
\end{itemize}

\begin{figure}
    \centering
    \includegraphics[width=0.4\textwidth]{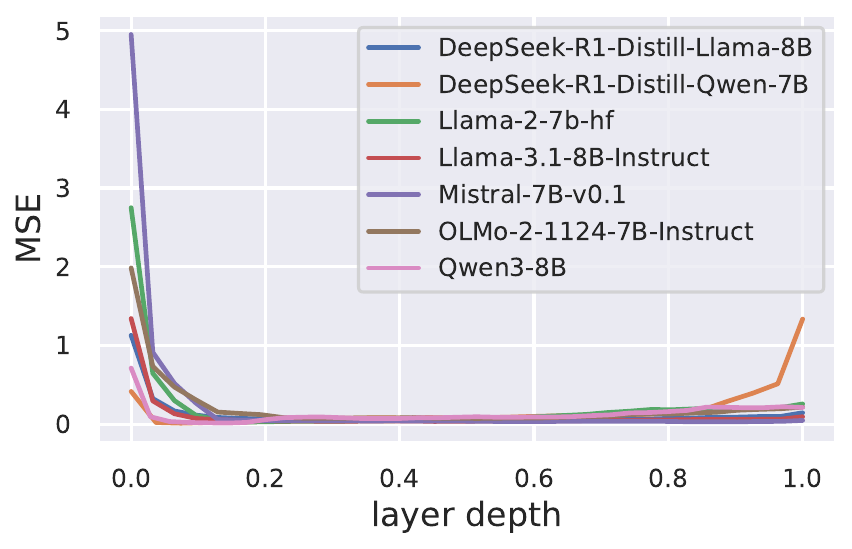}
    \caption{MSEs in log-space of regression probes on cross-notation data   of each LLM across layers.}
    \label{fig:regression_mse_layers}
\end{figure}

\begin{figure*}
    \centering
    \includegraphics[width=1.0\linewidth]{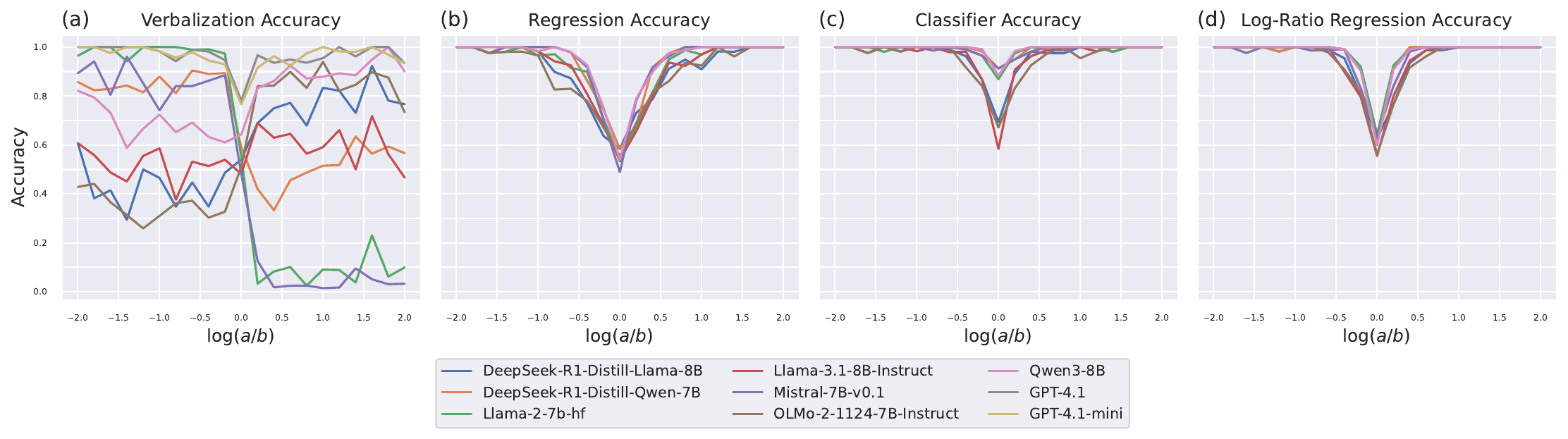}
    \caption{Accuracy of cross-notation comparison ($a \stackrel{?}{>} b$) versus the relative magnitude of the two numbers ($\log_2(a/b)$).  (a)~Verbalized comparison using one-shot prompting;  (b)~Comparison of the two values predicted by regression; (c)~Comparison via a classification probe; (d)~Comparison via the log-ratio predicted by regression.  Note that (b)--(d) require access to the hidden states, so they do not include the large closed-source models GPT-4.1 and GPT-4.1-mini.  See \cref{fig:log-ratio-plot-individual-models} for individual models' results.}
    \label{fig:log-ratio-plot}
\end{figure*}

\section{Experimental Setup}
\label{sec:experimental-setup}

\paragraph{Data}\label{experimental setup}
We construct a synthetic \texttt{cross-notation} dataset to support our studies of both regression and comparison.  
To study regression probing on realistic scientific text, we derive an \texttt{arXiv} dataset from the peS2o dataset \citep{peS2o}.  See  \cref{appendix:data-setup} for details of both datasets.%

In \texttt{cross-notation}, numeral pairs are constructed so that each pair consists of one numeral in scientific notation and the other in normal form (e.g., $5.7 \times 10^2$ vs.\@ $560$).\footnote{We found LLMs can compare numerals of the same notation with 100\% accuracy.}

For verbalization on \texttt{cross-notation}, we prompt with a one-shot demonstration since the model has not been trained for the task (see \cref{appendix:inference-setup} for more details):

{\small
\begin{verbatim}
Q: Which is larger, 9.9 × 10^2 or 100? A: 9.9 × 10^2
Q: Which is larger, {a} or {b}? A:
\end{verbatim}
}

\noindent In contrast, we train and test our \texttt{cross-notation} probes on a zero-shot prompt (no demonstration), to better isolate the model’s intrinsic encoding:

{\small
\begin{verbatim}
Q: Which is larger, {a} or {b}? A:
\end{verbatim}
}

\paragraph{Probing}
For each layer of the transformer LLM, we extract hidden states from the zero-shot prompt and train separate probes. Linear regression probes are trained on the last tokens of the numerals to predict their $\log_2$ magnitudes and the last token of the prompt to predict $\log_2(a/b)$. Linear classification probes (using logistic regression) are trained on the last token of the prompt to predict 
whether $\verb+{a}+ > \verb+{b}+$.  
See \cref{appendix:probing-setup} for details.

\paragraph{Finetuning}
To investigate whether internal magnitude representations can be leveraged to improve numerical reasoning, we incorporate auxiliary probing loss during finetuning on \texttt{cross-notation}. We add probe heads to a single layer and finetune on the standard language modeling loss {\em plus}
the classification probing loss,\footnote{We experimented with adding regression probing losses but found no improvement in performance (see \cref{appendix:finetuning}).}
$\mathcal{L}_{\text{total}} = \mathcal{L}_{\text{LM}} 
+ \beta \mathcal{L}_{\text{cls}}$,
where 
the hyperparameter $\beta$ controls the strength of auxiliary supervision. 
See \cref{appendix:finetuning} for details.

\paragraph{Models}
To ensure that the experimental results are not specific to a certain model's architecture or training procedure, we select a set of widely used open-weight 7B--8B LLMs, including base pretrained, instruction-tuned, and distilled models. We additionally evaluate GPT-4.1 and GPT-4.1-mini (OpenAI's non-reasoning frontier models) to assess whether verbalization difficulties persist in larger, more capable models.\jason{do they do well on math?}

\section{Experimental Results}

\subsection{Magnitude Probing}
\label{sec:regression-probing}
To investigate whether LLMs internally encode numerical magnitude information, we first train linear regression probes on hidden states to predict the $\log_2$ values of numerals. \cref{fig:scatter-all} shows scatterplots of predicted versus true log-magnitudes for Mistral-7B across different datasets and notation types.

The strong correlation observed in \cref{fig:scatter-all} suggests that, on the \texttt{cross-notation} dataset, log-magnitudes of decimals and scientific notation numerals are encoded linearly in the LLM's hidden representations. 
Even in mixed probing, where probes are trained and tested on both decimal and scientific notation tokens, linear regressors achieve high performance on \texttt{cross-notation} with $R^2 > 0.998$  (\cref{fig:scatter-mix-synthetic}).

On \texttt{arXiv} (\cref{fig:scatter-mix-arxiv}), the linear probe is not as precise, though it still achieves a correlation of $\rho=71\%$ ($R^2=0.51$).
Numbers in \texttt{arXiv} play many more different \emph{semantic roles} \citep[App. F]{lo-wang-2020-s2orc}.  Perhaps these are represented in different subspaces, so that a single linear probe is not powerful enough.

Even so, this latter model can predict magnitudes on held-out \texttt{arXiv} data to 19.06\% median relative error.  Thus, this layer of Mistral-7B contains a noisy invariant linear representation of each numeral's magnitude---which the next layer's attention mechanism could learn to extract through linear query/key/value projections, providing ``number sense''.  Note that the better 2.3\% median relative error on \texttt{cross-notation} (\cref{fig:scatter-mix-synthetic}) is possible only because the models are overfitted to very specific prompt formats.\footnote{These models are also underdetermined because they have only seen one kind of data.  They do not generalize well even within the \texttt{cross-notation} dataset: the \cref{fig:scatter-dec-synthetic} model performs much worse on the \cref{fig:scatter-sci-synthetic} data ($R^2=0.56$), and vice-versa ($R^2=0.34$).}

\subsection{Comparison Probing}\label{sec:classification-probing}

Despite the regression probes' low error in log-space, their predictions are too noisy to compare close-by values.
In \cref{fig:log-ratio-plot}b, we use the regression probe's predicted magnitudes to compare the two numbers in the prompt.  
We observe across all models that this method degrades when the absolute log$_2$-ratio of the two numbers drops below 1 (i.e., when the two numbers differ by less than a factor of 2). When $|\log_2(a/b)|<0.2$, the regression probes' predictions are too noisy to meaningfully compare the numbers, resulting in random chance accuracy.\looseness=-1

\begin{figure}[t]
    \centering
    \includegraphics[width=0.4\textwidth]{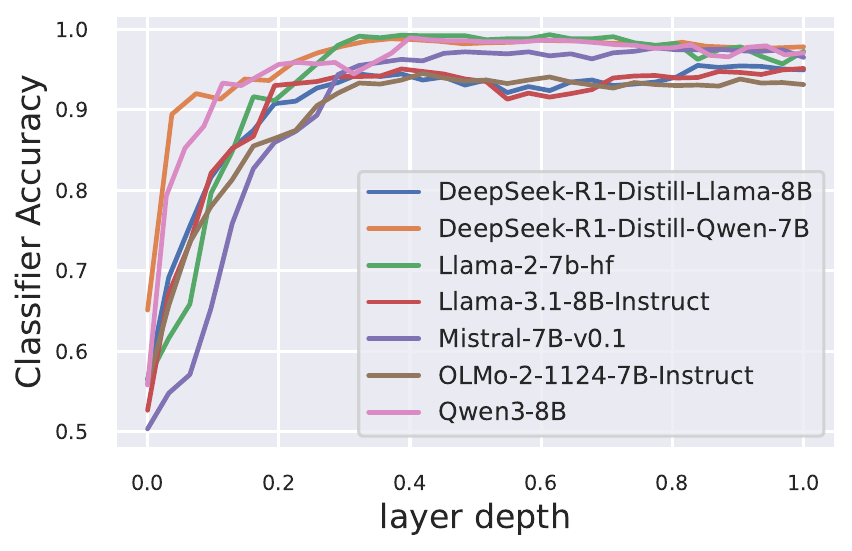}
    \caption{Logistic classifier accuracies on \texttt{cross-notation} of each LLM across layers.}
    \label{fig:classification_layers}
\end{figure}

However, the LLM's internal states also seem to include a comparison of the two numbers.  We additionally train a binary classifier on the \texttt{cross-notation} dataset.
\cref{fig:classification_layers} plots accuracy as a function of layer depth.
Higher layers consistently exceed 95\% accuracy.
\cref{fig:log-ratio-plot}c shows the accuracy with respect to the log-ratio of two numbers.
We observe that, while classifier probes also suffer when the two numbers are close, they generally perform better than regression-based comparison---and much better for 4 of the 7 open-source models.

The internal states encode not only which number is bigger, but how much bigger it is.  We trained additional regression probes to predict $\log_2(a/b)$ from the last hidden state of the prompt.  These achieved overall predictive accuracy similar to the regression probes of \cref{sec:regression-probing} (see \cref{fig:log-ratio-correlation-regression-all} in \cref{sec:additional-probing-results}).  
Their accuracy may still degrade when $a\approx b$, but as $a$ approaches $b$, comparison based on predicted $\log_2(a/b) > 0$ (\cref{fig:log-ratio-plot}d) does not degrade quite as quickly as comparison based on predicted $\log_2(a) >$ predicted $\log_2(b)$ (\cref{fig:log-ratio-plot}b).
\Cref{fig:log-ratio-plot-individual-models} clearly shows that for each model, the log-ratio comparison method is more accurate at all $a/b$ values.  Indeed, it behaves similarly to direct binary classification (\cref{fig:log-ratio-plot}c)---except for the 4 models whose binary classifiers are much more robust.  Presumably those models are internally performing the actual comparison task requested by our prompt.

\subsection{Verbalization}
\label{sec:verbalization-results}

\cref{tab:verbal vs. probing} shows each LLM’s verbalized response on \texttt{cross-notation} using a one-shot prompt.
Across a variety of 7B--8B open-weight models, the best model reaches only 70\% accuracy; the worst is at chance (50\%).
While few-shot prompting improves performance (cf. \cref{sec:additional-error-analysis}), we emphasize one-shot performance because numeracy \emph{should be} an innate ability of LLMs and such basic questions should require as few examples as possible.

To visualize the pattern of errors, \cref{fig:log-ratio-plot}a shows the verbalized comparison accuracy as $\log_2(a/b)$ changes. 
Though GPT-4.1 and GPT-4.1-mini achieve over 92\% accuracy overall, they have errors at all ratios and especially when the two numbers are close.
Interestingly, we observe that Llama-2-7B and Mistral-7B disregard the values of the numbers and almost always answer with the second number. We conduct further experiments in \cref{sec:additional-error-analysis} and show that, in the one-shot setting, those two models lack awareness of the values of numbers and superficially imitate the example in the one-shot prompt.  We present few-shot results and a more thorough error analysis in \cref{sec:additional-error-analysis}. 

\begin{table*}[ht]
\centering
\small
\begin{tabular}{lrrrrr}
\toprule
\textbf{Model} &
\textbf{Base} &
\textbf{Finetuned} &
\textbf{\shortstack{Error Rate \\ Reduction}} &
\textbf{\shortstack{Finetuned \\ with Probing Loss}} &
\textbf{\shortstack{Further Error \\ Rate Reduction}} 
\\
\midrule
DeepSeek-R1-Distill-Llama-8B &
57.81 &
86.69 &
68.5\% & 
89.69 &
22.5\%  
\\
DeepSeek-R1-Distill-Qwen-7B &
64.19 & 
84.06 &
55.5\% & 
95.19 &
69.8\%  
\\
Llama-2-7B &
50.81 &
94.44 &
88.7\% & 
94.56 &
2.2\%   
\\
Llama-3.1-8B-Instruct &
55.06 &
96.44 &
92.1\% & 
95.81 &
$-$17.7\%   
\\
Mistral-7B-v0.1 &                   
50.00 &
95.69 &
91.4\% & 
99.31 &
84.0\%   
\\
OLMo-2-1124-7B &                   
53.44 &
87.75 &
73.7\% & 
90.25 &
20.4\%   
\\
Qwen3-8B &           
70.00 &
96.19 &
87.3\% & 
99.00 &
73.8\%   
\\ \midrule
Average &
57.33 &
91.61 &
80.3\% & 
94.83 &
38.4\%   
\\
\bottomrule
\end{tabular} 
\caption{Cross-notation comparison accuracy of verbalized responses.  Accuracy numbers are percentages without the \% sign. Most errors of the base model are fixed by finetuning.  Incorporating probing loss into the finetuning objective often fixes many of the \emph{remaining} errors (more than $\frac{2}{3}$ of them for 3 of the models). 
}
\label{tab:finetune-accuracy}
\vspace{-\baselineskip}
\end{table*}

\subsection{Correlation}\label{sec:correlation}

\begin{figure}[t]
    \centering
    \includegraphics[width=0.44\textwidth]{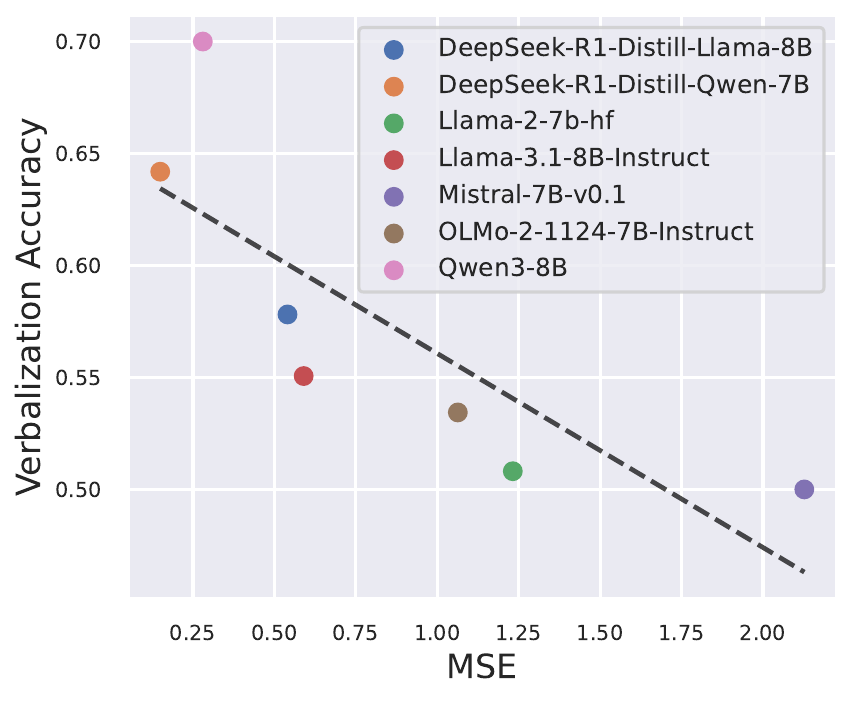}
    \caption{Average performance of linear regression probes at the first 3 layers correlates with model's verbalization accuracy on \texttt{cross-notation.}}
    \label{fig:correlation-regression-mse}
\end{figure}

\begin{figure}[t]
    \centering
    \includegraphics[width=0.4\textwidth]{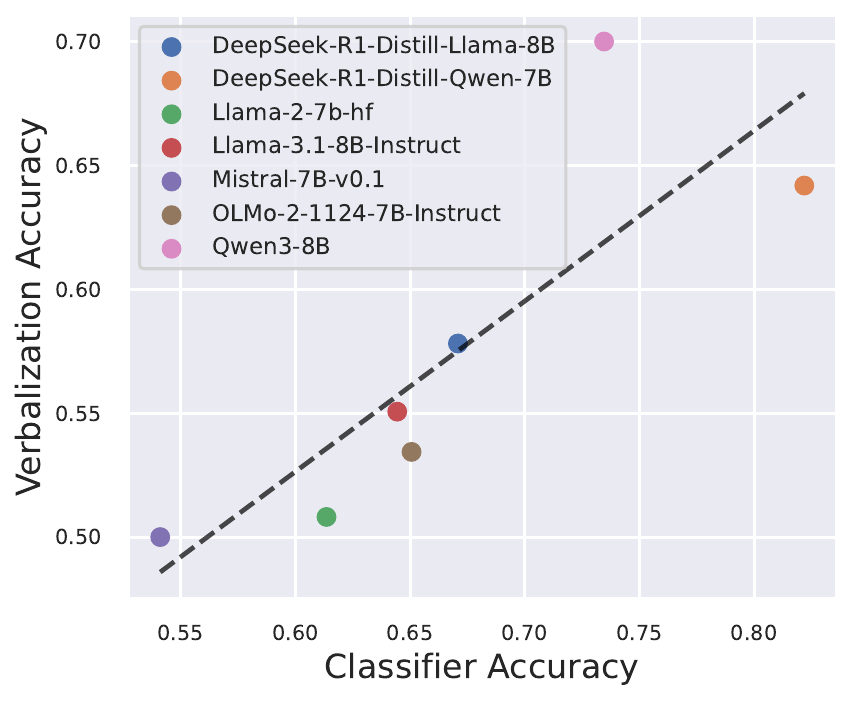}
    \caption{Average performance of logistic classification probes at the first 3 layers predicts the model's verbalization accuracy on \texttt{cross-notation.}}
    \label{fig:correlation-classification}
\vspace{-\baselineskip}
\end{figure}

To explore how internal representations relate to verbalized performance, we investigate the correlation between probe performance and verbal accuracy across models.

We observe that for scientific notation numerals, the best probes (\crefrange{fig:scatter-sci-synthetic}{fig:scatter-mix-synthetic}) are near the top of the transformer.  These representations presumably arrive too late in processing to support the comparisons required by the verbalization task. We further observe that 
representations that support accurate linear regression are only available after several layers (\cref{fig:regression_mse_layers}), and representations that support accurate classification arrive even later (\cref{fig:classification_layers}).  

We speculate that earlier representations are more useful for downstream tasks. 
Indeed, \cref{fig:correlation-regression-mse,fig:correlation-classification} reveal a strong correlation \emph{across diverse 7B--8B models} between the average performance of probes from the first three layers (even though that performance is still low) and verbalized comparison accuracy on the \texttt{cross-notation} dataset. 

This correlation suggests that higher quality of internal numerical representations is linked with improved verbalization performance.

\subsection{Finetuning}\label{sec:finetuning}

To test whether the link is causal, we incorporate probing loss as an auxiliary objective during finetuning, i.e., we jointly train the LLM and the probe.
\cref{tab:finetune-accuracy} compares three conditions: base model, standard finetuning ($\beta=0$), and finetuning with probing loss ($\beta=0.02$).\footnote{See hyperparameter tuning in \cref{appendix:finetuning}.}
Through hyperparameter tuning, we select the probe layer depth to be 90\%.
$\beta>0$ encourages the representations to better support the comparison task that will be needed by verbalization.  Through backpropagation, this affects the representations at earlier layers as well. 

Ordinary finetuning fixes $> \frac{4}{5}$ of the total errors made by these 7 models (in the sense of reducing error rate).  But augmenting the finetuning with probing loss fixes $> \frac{1}{3}$ of the total \emph{remaining} errors.  For the 3 Mistral- and Qwen-based models, augmentation fixes $> \frac{2}{3}$ of remaining errors, bringing 2 of them to $\geq 99$\% accuracy.  This suggests the possibility of improving LLMs' verbal numeracy by explicitly improving their internal representations.

\section{Conclusion}

We study a fundamental question in LLM numeracy (see \cref{sec:related} for related work): Do language models know how big numbers are? 
We found a striking fact: While LLMs do have strong internal numerical representations that can be exposed by both regression and classification probes, they struggle to answer cross-notation comparisons, performing only marginally above chance.\jason{Yet---as our title says---their internal representations \emph{do} successfully perform cross-notation comparison.}

We demonstrate that early-layer representations strongly predict verbalized performance, and auxiliary probing loss enhances finetuning with an additional 3.22\% improvement.\jason{does this improvement transfer to regression (when you retrain the probes)?  Possibly even on arXiv? \response{fengting}after finetuning with probing losses, the losses of probe heads are certainly decreasing, but by retraining the probes there's no immprovement.}
These results suggest that LLMs possess sophisticated numerical knowledge they cannot reliably access during generation unless fine-tuned for a task, but also that better knowledge causally leads to better generation.

Our work highlights both limitations and opportunities.  Current LLMs know more about numbers 
than they can say;
and targeted tuning on these two goals (representation, verbalization) is synergistic.

\section*{Limitations}

\begin{itemize}
    \item \textbf{Probing reveals representation, not usage.} 
    Unlike causal tracing approaches \citep{lindsey2025biology}, we did not investigate how the representations are actually used during inference.  Our intervention trained the model to produce both better representations ($\mathcal{L}_{\text{cls}}$) and better text output ($\mathcal{L}_{\text{LM}}$).  We found a synergy between these two objectives, but we did not intervene on the representations directly at inference time to see how this changed the text output. 
    \item \textbf{Synthetic-real gap.} Most of our core findings—such as probing performance, verbalization failure, and probe-based finetuning gains—are demonstrated on controlled synthetic data.  We do not systematically validate these results on real scientific text such as arXiv papers. The extent to which our conclusions generalize to such text (and to other aspects of numeracy) remains to be established.
    \item \textbf{Linear probes may not be enough.}\jason{check that all of the ideas from reviewer-author discussion on OpenReview were incorporated} Our tests on arXiv showed that a single linear probe will not necessarily be enough to work beyond our simple synthetic data setting.  On real data, we might try using a two-layer neural network, or using the minimum loss of several linear probes that look at different subspaces.  Or perhaps these nonlinearities are unnecessary if through finetuning a linear probe, we can successfully change the internal representations so that they do support linear probing after all, even on arXiv.  That is, finetuning might recruit a dimension to encode numeric log-magnitude, which might improve the model's abilities in downstream verbal numeracy tasks.  We leave these questions to future work.
    \item \textbf{Probing findings are limited to 7B--8B LLMs.} Our probing results are limited to 7B--8B open-weight models where we can access internal representations.
    We cannot probe GPT-4.1 and GPT-4.1-mini, so it remains unknown whether larger models encode numerical information similarly, or whether the links between internal numeracy and verbal performance are preserved. 
    \item \textbf{Limited diversity in test examples.}
    Our synthetic cross-notation dataset has several constraints. First, all exponents are positive, but negative exponents like $10^{-3}$ are common in scientific literature.
    Second, our few-shot demonstration examples (\cref{appendix:inference-setup}) did not include examples where numbers are very close (e.g., 1234 vs.\@ $1.241 \times 10^3$); perhaps targeted demonstrations on such cases would reduce their error rate. Third, the same prompt template and few-shot examples are used for all test problems, so our results may depend on those specific ordering and selection patterns.
    In retrospect, we should have randomly varied the prompt format and few-shot examples within the test set.

    \item \textbf{Semantic context is not considered.} Our probing methods focus on the magnitudes of numerals in isolation, without considering context such as units. In practice, numeracy requires contextual understanding—for instance, 5 miles is greater than 50 meters, even though $5 < 50$, and 50 meters and 50 kg are incomparable. While LLMs likely represent such contextual information internally, we leave this aspect for future work.
\end{itemize}

\section*{Acknowledgments}

This work was supported by DOE Award No.\@ DE-SC0025653 from the U.S. Department of Energy.  Johns Hopkins University provided use of the Rockfish compute cluster (\cref{appendix:cluster}).  We thank Patrick Emami, Jared Willard, and members of the Argo Lab for questions and feedback during the project.

\bibliography{acl_latex}

\begin{thebibliography}{18}
\providecommand{\natexlab}[1]{#1}

\bibitem[{Hanna et~al.(2023)Hanna, Liu, and Variengien}]{hanna2023how}
Michael Hanna, Ollie Liu, and Alexandre Variengien. 2023.
\newblock \href {https://openreview.net/forum?id=p4PckNQR8k} {How does {GPT}-2
  compute greater-than?: Interpreting mathematical abilities in a pre-trained
  language model}.
\newblock In \emph{Thirty-seventh Conference on Neural Information Processing
  Systems}.

\bibitem[{Hu et~al.(2022)Hu, Shen, Wallis, Allen-Zhu, Li, Wang, Wang, and
  Chen}]{hu2022lora}
Edward~J Hu, Yelong Shen, Phillip Wallis, Zeyuan Allen-Zhu, Yuanzhi Li, Shean
  Wang, Lu~Wang, and Weizhu Chen. 2022.
\newblock \href {https://openreview.net/forum?id=nZeVKeeFYf9} {Lo{RA}: Low-rank
  adaptation of large language models}.
\newblock In \emph{International Conference on Learning Representations}.

\bibitem[{Jiang et~al.(2023)Jiang, Sablayrolles, Mensch, Bamford, Chaplot,
  de~las Casas, Bressand, Lengyel, Lample, Saulnier, Lavaud, Lachaux, Stock,
  Scao, Lavril, Wang, Lacroix, and Sayed}]{jiang2023mistral7b}
Albert~Q. Jiang, Alexandre Sablayrolles, Arthur Mensch, Chris Bamford,
  Devendra~Singh Chaplot, Diego de~las Casas, Florian Bressand, Gianna Lengyel,
  Guillaume Lample, Lucile Saulnier, Lélio~Renard Lavaud, Marie-Anne Lachaux,
  Pierre Stock, Teven~Le Scao, Thibaut Lavril, Thomas Wang, Timothée Lacroix,
  and William~El Sayed. 2023.
\newblock \href {https://arxiv.org/abs/2310.06825} {Mistral {7B}}.
\newblock \emph{Preprint}, arXiv:2310.06825.

\bibitem[{Jimenez et~al.(2024)Jimenez, Yang, Wettig, Yao, Pei, Press, and
  Narasimhan}]{jimenez2024swebench}
Carlos~E Jimenez, John Yang, Alexander Wettig, Shunyu Yao, Kexin Pei, Ofir
  Press, and Karthik~R Narasimhan. 2024.
\newblock \href {https://openreview.net/forum?id=VTF8yNQM66} {{SWE}-bench: Can
  language models resolve real-world github issues?}
\newblock In \emph{The Twelfth International Conference on Learning
  Representations}.

\bibitem[{Li et~al.(2025)Li, Chen, XU, Li, Hu, Teng, Li, Qiu, Zhang, Li, and
  Chen}]{li2025exposingnumeracygapsbenchmark}
Haoyang Li, Xuejia Chen, Zhanchao XU, Darian Li, Nicole Hu, Fei Teng, Yiming
  Li, Luyu Qiu, Chen~Jason Zhang, Qing Li, and Lei Chen. 2025.
\newblock \href {https://arxiv.org/abs/2502.11075} {Exposing numeracy gaps: A
  benchmark to evaluate fundamental numerical abilities in large language
  models}.
\newblock \emph{Preprint}, arXiv:2502.11075.

\bibitem[{Lindsey et~al.(2025)Lindsey, Gurnee, Ameisen, Chen, Pearce, Turner,
  Citro, Abrahams, Carter, Hosmer, Marcus, Sklar, Templeton, Bricken,
  McDougall, Cunningham, Henighan, Jermyn, Jones, Persic, Qi, Thompson,
  Zimmerman, Rivoire, Conerly, Olah, and Batson}]{lindsey2025biology}
Jack Lindsey, Wes Gurnee, Emmanuel Ameisen, Brian Chen, Adam Pearce,
  Nicholas~L. Turner, Craig Citro, David Abrahams, Shan Carter, Basil Hosmer,
  Jonathan Marcus, Michael Sklar, Adly Templeton, Trenton Bricken, Callum
  McDougall, Hoagy Cunningham, Thomas Henighan, Adam Jermyn, Andy Jones, and 8
  others. 2025.
\newblock \href
  {https://transformer-circuits.pub/2025/attribution-graphs/biology.html} {On
  the biology of a large language model}.
\newblock \emph{Transformer Circuits Thread}.

\bibitem[{Lo et~al.(2020)Lo, Wang, Neumann, Kinney, and
  Weld}]{lo-wang-2020-s2orc}
Kyle Lo, Lucy~Lu Wang, Mark Neumann, Rodney Kinney, and Daniel Weld. 2020.
\newblock \href {https://doi.org/10.18653/v1/2020.acl-main.447} {{S}2{ORC}: The
  semantic scholar open research corpus}.
\newblock In \emph{Proceedings of the 58th Annual Meeting of the Association
  for Computational Linguistics}, pages 4969--4983, Online. Association for
  Computational Linguistics.

\bibitem[{Loshchilov and
  Hutter(2019)}]{loshchilov2019decoupledweightdecayregularization}
Ilya Loshchilov and Frank Hutter. 2019.
\newblock \href {https://arxiv.org/abs/1711.05101} {Decoupled weight decay
  regularization}.
\newblock \emph{Preprint}, arXiv:1711.05101.

\bibitem[{Soldaini and Lo(2023)}]{peS2o}
Luca Soldaini and Kyle Lo. 2023.
\newblock {peS2o (Pretraining Efficiently on S2ORC) Dataset}.
\newblock Technical report, {Allen Institute for AI}.
\newblock ODC-By, \url{https://github.com/allenai/pes2o}.

\bibitem[{Song and Bahri(2025)}]{decoding_regression}
Xingyou Song and Dara Bahri. 2025.
\newblock \href {https://arxiv.org/abs/2501.19383} {Decoding-based regression}.
\newblock \emph{Computing Research Repository}, arXiv:2501.19383.

\bibitem[{Stolfo et~al.(2023)Stolfo, Belinkov, and Sachan}]{stolfo2023a}
Alessandro Stolfo, Yonatan Belinkov, and Mrinmaya Sachan. 2023.
\newblock \href {https://openreview.net/forum?id=aB3Hwh4UzP} {A mechanistic
  interpretation of arithmetic reasoning in language models using causal
  mediation analysis}.
\newblock In \emph{Proceedings of the 2023 Conference on Empirical Methods in
  Natural Language Processing (EMNLP)}.

\bibitem[{Tang et~al.(2025)Tang, Yang, and Song}]{tang2025understanding}
Eric Tang, Bangding Yang, and Xingyou Song. 2025.
\newblock \href {https://openreview.net/forum?id=Wt6Iz5XNIO} {Understanding
  {LLM} embeddings for regression}.
\newblock \emph{Transactions on Machine Learning Research}.

\bibitem[{Taylor et~al.(2022)Taylor, Kardas, Cucurull, Scialom, Hartshorn,
  Saravia, Poulton, Kerkez, and
  Stojnic}]{taylor2022galacticalargelanguagemodel}
Ross Taylor, Marcin Kardas, Guillem Cucurull, Thomas Scialom, Anthony
  Hartshorn, Elvis Saravia, Andrew Poulton, Viktor Kerkez, and Robert Stojnic.
  2022.
\newblock \href {https://arxiv.org/abs/2211.09085} {Galactica: A large language
  model for science}.
\newblock \emph{Preprint}, arXiv:2211.09085.

\bibitem[{Wallace et~al.(2019)Wallace, Wang, Li, Singh, and
  Gardner}]{wallace-etal-2019-nlp}
Eric Wallace, Yizhong Wang, Sujian Li, Sameer Singh, and Matt Gardner. 2019.
\newblock \href {https://doi.org/10.18653/v1/D19-1534} {Do {NLP} models know
  numbers? {P}robing numeracy in embeddings}.
\newblock In \emph{Proceedings of the 2019 Conference on Empirical Methods in
  Natural Language Processing and the 9th International Joint Conference on
  Natural Language Processing (EMNLP-IJCNLP)}, pages 5307--5315, Hong Kong,
  China. Association for Computational Linguistics.

\bibitem[{Wu et~al.(2023)Wu, Irsoy, Lu, Dabravolski, Dredze, Gehrmann,
  Kambadur, Rosenberg, and Mann}]{wu2023bloomberggptlargelanguagemodel}
Shijie Wu, Ozan Irsoy, Steven Lu, Vadim Dabravolski, Mark Dredze, Sebastian
  Gehrmann, Prabhanjan Kambadur, David Rosenberg, and Gideon Mann. 2023.
\newblock \href {https://arxiv.org/abs/2303.17564} {{BloombergGPT}: A large
  language model for finance}.
\newblock \emph{Preprint}, arXiv:2303.17564.

\bibitem[{Yang et~al.(2025)Yang, Hu, Kang, Lin, and Zhang}]{yang2025number}
Haotong Yang, Yi~Hu, Shijia Kang, Zhouchen Lin, and Muhan Zhang. 2025.
\newblock \href {https://openreview.net/forum?id=BWS5gVjgeY} {Number cookbook:
  Number understanding of language models and how to improve it}.
\newblock In \emph{The Thirteenth International Conference on Learning
  Representations}.

\bibitem[{Zhang et~al.(2025)Zhang, Ding, Lv, Wang, Yin, Zhang, Yu, Wang, Li,
  Xiang, Zhuang, Wang, Qin, Zhang, Zhang, Cui, Xu, Chen, Fan, Xing, and
  Chen}]{zhang2025scientific}
Qiang Zhang, Keyan Ding, Tianwen Lv, Xinda Wang, Qingyu Yin, Yiwen Zhang, Jing
  Yu, Yuhao Wang, Xiaotong Li, Zhuoyi Xiang, Xiang Zhuang, Zeyuan Wang, Ming
  Qin, Mengyao Zhang, Jinlu Zhang, Jiyu Cui, Renjun Xu, Hongyang Chen, Xiaohui
  Fan, and 2 others. 2025.
\newblock \href {https://doi.org/10.1145/3715318} {Scientific large language
  models: A survey on biological \& chemical domains}.
\newblock \emph{ACM Comput. Surv.}, 57(6).

\bibitem[{Zhu et~al.(2025)Zhu, Dai, and Sui}]{zhu-etal-2025-language}
Fangwei Zhu, Damai Dai, and Zhifang Sui. 2025.
\newblock \href {https://aclanthology.org/2025.coling-main.47/} {Language
  models encode the value of numbers linearly}.
\newblock In \emph{Proceedings of the 31st International Conference on
  Computational Linguistics}, pages 693--709, Abu Dhabi, UAE.

\end{thebibliography}

\clearpage
\appendix
\onecolumn
\section{Related Work}\label{sec:related}\jason{is there room to move this into main paper?\response{fengting}5 page limit for camera-ready. Maybe no room for camera-ready, but we can do it for arxiv}
There is rich recent literature on internal numerical representations in large language models.
\Citet{wallace-etal-2019-nlp} demonstrate that numerical information can be extracted from contextual or static word embeddings.  
\citet{stolfo2023a} and \citet{hanna2023how} study internal mechanisms of LLMs' mathematical abilities.
Closely related to our work is \citet{zhu-etal-2025-language}, who provide evidence that language models encode the value of integers linearly in log-space.  
Most recently, \citet{lindsey2025biology} attempted to recover the internal procedure through which a specific LLM performs additions of two-digit numbers.

Our work also overlaps with recent LLM numeracy benchmark efforts.  
Recently, \citet{yang2025number} designed a comprehensive benchmark for evaluating LLMs' black-box responses to basic synthetic numerical questions.  However, they did not investigate internal numerical representations.  
Similarly, \citet{li2025exposingnumeracygapsbenchmark} designed a basic numerical benchmark based on real-world datasets, using a similar blackbox approach.
In a similar effort to extract precise numbers from LLM encodings, \citet{tang2025understanding} and \citet{decoding_regression} studied whether LLM encodings can be used to tackle classical regression tasks.

\section{Data Setup}
\label{appendix:data-setup}
\subsection{Synthetic Data}
\label{appendix:synthetic-data-setup}
The \texttt{cross-notation} dataset in \cref{experimental setup} contains 11,200 comparison problems, with 8,000 samples for training, 1,600 for validation and 1,600 for evaluation.
We generate 1,400 numeral pairs for each digit length between 2 and 9 digits, randomly converting one numeral in each pair to scientific notation while keeping the other in standard notation.

We construct two variants of this dataset: \texttt{int-sci} (integer vs.\@ scientific notation) and \texttt{dec-sci} (decimal vs.\@ scientific notation). 
In \texttt{int-sci}, both numbers start as integers before one is converted to scientific notation (e.g., $570$ vs.\@ $580$ $\rightarrow$ $5.7 \times 10^2$ vs.\@ $580$). In \texttt{dec-sci}, we first append 0--4 random decimal digits independently to each number before conversion (e.g., $570.23$ vs.\@ $871.6$ $\rightarrow$ $570.23$ vs.\@ $8.716 \times 10^2$).\jason{is it always the same number of decimal digits for both numbers, as in this example?  And do we append a decimal point even when the number of digits is 0?  That is actually important for where we probe.\response{fengting} no, we generate numbers of decimal digits for a and b separately. you probably mistook $8.716 \times 10^2$ as 87.16 (corrected); when the number of decimal digits is 0, we still append .0 to the number. For example, the integer part of a is 342, and the number of decimal digits is 0, then we have 342.0 as the decimal. I added this to the paragraph}
When zero decimal digits are selected, we append ``.0'' to maintain consistent formatting (e.g., $342.0$).
These two datasets are functionally interchangeable for our experiments. We report results using \texttt{dec-sci} in \cref{fig:log-ratio-plot,fig:log-ratio-plot-individual-models,fig:alt-prompt,fig:digit-plot,fig:magnitude-plot}, with \texttt{int-sci} used for all other experiments on \texttt{cross-notation}.

\subsection{Scientific Data}
We utilize the \texttt{peS2o} dataset \cite{peS2o}, which contains approximately 40 million open-access academic papers from arXiv that have been cleaned, filtered, and formatted for language model pre-training. \texttt{peS2o} consists of two subsets: \texttt{s2orc}, which provides full-text papers \cite{lo-wang-2020-s2orc}\jason{i moved this citation here from the intro: is it appropriate?\response{fengting}it's appropriate here}, and \texttt{s2ag}, which contains titles and abstracts only. 

To build \texttt{arXiv} data, we only use the \texttt{s2orc} subset. 
We first shuffle the corpus using \texttt{terashuf} and select the first 100k articles from \texttt{s2orc}. We then employ regular expressions to extract numerals in standard decimal notation (\verb|(?<!\d\.)\d+\.\d+(?!\.\d)|) and in scientific notation (\verb|(?:\d+(?:\.\d+)?)\s*×\s*10\s+[-+]?\d+|).
To avoid overflowing GPU memory, we filter out papers exceeding 30,000 tokens.
We process complete documents through the language models and extract hidden states at the final token position of each identified numeral. The resulting processed \texttt{arXiv} dataset contains 5,000 samples, divided into 4,000 training samples and 500 validation samples and 500 evaluation samples.

\section{Inference and Prompting Setup}
\label{appendix:inference-setup}

For verbalization tasks, all models are evaluated with temperature 0 (greedy decoding) to ensure deterministic and reproducible outputs.
\subsection{Open-weight Models}
For the 7B--8B open-weight models, we use the Hugging Face \texttt{Transformers} library for decoding. 
Prompts are formatted as plain text with ``Q:'' and ``A:'' markers.
The one-shot demonstration for \texttt{int-sci} is shown in \cref{sec:experimental-setup}. For \texttt{dec-sci}, we use:
{
\begin{verbatim}
Q: Which is larger, 9.9 × 10^2 or 899.9? A: 9.9 × 10^2
Q: Which is larger, {a} or {b}? A:
\end{verbatim}
}

For $k$-shot experiments using \texttt{int-sci} (\cref{sec:additional-error-analysis}), where $k \in [1,5]$, we prepend the first $k$ of these examples:
{
\begin{verbatim}
Q: Which is larger, 9.9 × 10^2 or 100? A: 9.9 × 10^2
Q: Which is larger, 161230 or 7.182 × 10^5? A: 7.182 × 10^5
Q: Which is larger, 713 or 4.78 × 10^2? A: 713
Q: Which is larger, 1.354 × 10^6 or 4906723? A: 4906723
Q: Which is larger, 20834 or 6.5 × 10^3? A: 20834
\end{verbatim}
}
\noindent followed by the usual test question:
{
\begin{verbatim}
Q: Which is larger, {a} or {b}? A:
\end{verbatim}
}

\subsection{GPT Models}
GPT-4.1 and GPT-4.1-mini are called via the OpenAI Chat Completions API with the system prompt ``Just answer with a number.''
The questions and answers are provided as messages with the \texttt{user} and \texttt{assistant} roles respectively, rather than being marked with \texttt{Q:} and \texttt{A:} strings.
The content of the few-shot examples is identical to that above.

\section{Probing Setup}
\label{appendix:probing-setup}

\subsection{Obtaining Hidden States}
We feed each input into the LLMs and extract the hidden states at every layer. For each layer, we identify specific token positions corresponding to our target representations and save their hidden states $\bm{H} \in \mathbb{R}^{n \times d_{\text{model}}}$ for subsequent probing experiments, where $n$ is the number of samples and $d_{\text{model}}$ is the hidden dimension.

\subsection{Training Regression Probes}
For regression probes, we extract hidden states from the last token of each numeral. 
Given hidden states $\bm{H}$ and their corresponding numerical values $\bm{x}\in \mathbb{R}^{n}$, we train a linear regressor $\mathcal{R}$ that predicts $\bm{y} = \bm{H}\bm{w} + b\,\bm{1}$, where $\bm{w} \in \mathbb{R}^{d_{\text{model}}}$ and $b\in\mathbb{R}$ are the learned parameters.

We apply $\log_2$ transformation to compress the numerical range and emphasize relative magnitude relationships, using an $\ell_2$ regularizer ($\lambda=1$) for training (i.e., ridge regression):
\begin{equation*}
   \bm{w}^*, b^* = \mathop{\arg\min}\limits_{\bm{w}, b} ||\log_{2}(\bm{x})-\bm{H}\bm{w}-b\,\bm{1}||^2_2 + \lambda ||\bm{w}||^2_2
\end{equation*}
The trained probe predicts logarithmic magnitudes as $\bm{y} = \bm{H}\bm{w}^* + b^*\,\bm{1}$.

\subsection{Training Classification Probes}
For classification probes, we extract hidden states from the last token of the entire prompt after processing both numerals. We frame the task as binary classification to predict which numeral is larger in a given pair $(a, b)$.

Given hidden states $\bm{H} = \{\bm{h}_1,\ldots,\bm{h}_n\} \in \mathbb{R}^{n \times d_{\text{model}}}$ and binary labels $\mathbf{y} \in \{0, 1\}^{n}$ (where $y_i = 1$ if the first numeral is larger), we train a logistic regression model with an $\ell_2$ regularizer ($\gamma=1$):
\begin{align*}
&\bm{w}^*, b^* = \mathop{\arg\max}\limits_{\bm{w}, b} \Bigg\{ \sum_{i=1}^{n} y_i \log(\sigma(\bm{h}_i^\top\bm{w} + b)) + (1-y_i) \log(1-\sigma(\bm{h}_i^\top\bm{w} + b)) 
- \frac{1}{2\gamma} \lVert \bm{w} \rVert^2 \Bigg\}
\end{align*}
where $\sigma$ is the sigmoid function. Given a hidden state $\bm{h}\in\mathbb{R}^{d_\text{model}}$, the probe estimates $P(a > b) = \sigma(\bm{h}^\top\bm{w}^*  + b^*)$.  Our binary classifier predicts $a > b$ iff this probability exceeds $\frac{1}{2}$, i.e., iff $\bm{h}^\top\bm{w}^* + b^* > 0$.

\section{Finetuning Setup}
\label{appendix:finetuning}
The total loss is initially comprised of three terms:
\begin{equation*}
\mathcal{L}_{\text{total}} = \mathcal{L}_{\text{LM}} 
 + \alpha \mathcal{L}_{\text{reg}} 
+ \beta \mathcal{L}_{\text{cls}}
\end{equation*}
We finetune on the training set from \cref{appendix:synthetic-data-setup} using LoRA \citep{hu2022lora} with settings \texttt{lora\_r}=16, \texttt{lora\_alpha}=32, \texttt{lora\_dropout}=0.1, \texttt{target\_modules}=\texttt{[q\_proj, v\_proj]}. 
We use AdamW \citep{loshchilov2019decoupledweightdecayregularization} as our optimizer, finetuning for 3 epochs with batch size 16.

We first initialize the probe parameters by fitting them in our usual way (\cref{appendix:probing-setup}) on the same training data, before proceeding with finetuning of all parameters as above.

Using the validation set from \cref{appendix:synthetic-data-setup}, we perform grid-search on probe and finetuning hyperparameters: $\alpha,\beta \in {\{0, 0.01, 0.02, 0.05, 0.1, 0.2, 0.5, 1, 2, 5, 10, 20, 50, 100\}}$, learning rate $\in$ \texttt{\{2e-6, 5e-6, 1e-5, 2e-5, 5e-5, 1e-4, 2e-4\}}, and probing layer depth $\in$ \texttt{\{10\%, 20\%, 30\%, 40\%, 50\%, 60\%, 70\%, 80\%, 90\%, 100\%\}}.  We select the hyperparameters that achieve the highest verbalization accuracy on validation data.

These optimal hyperparameters are $\alpha=0$, $\beta=0.02$, learning rate = 5e-5, probing layer depth = 90\%. Thus the regression probing loss term is omitted in \cref{sec:experimental-setup} since including it is suboptimal.  In fact, we found that $\alpha > 0$ was suboptimal even when $\beta=0$, indicating that attempting to improve regression loss did not help with the verbalized comparison.  Presumably that was because regression on the \texttt{cross-notation} dataset was already extremely accurate.  On a dataset like \texttt{arXiv}, however,  there would be more room to improve regression, and this might benefit the downstream task of verbalized comparison.

\section{Evaluation Metrics}
\label{appendix:metrics}

\textbf{Mean squared error (MSE)} is the average squared difference between probe predictions and actual values in log space.

\noindent \textbf{Relative error (RE)} (reported in our paper abstract) converts the median absolute error in log space to a relative error as shown below. 

\noindent \textbf{Approximate accuracy (AAcc)} evaluates whether the predicted number is approximately the same as the original number in normal (non-log) space, namely with an error margin of $<1\%$. 
Higher AAcc indicates that the number encoding is more likely to be precise.

\begin{align}
    \mathrm{MSE}(\bm{y}, \bm{x}) &= \mathrm{mean}((\bm{y}-\log_{2}\bm{x})^2)
    \\
    \mathrm{RE}(\bm{y}, \bm{x}) &= 2^{\mathrm{median}(|\bm{y}-\log_{2}\bm{x}|)} - 1 = \mathrm{median}\left(\max\left(\frac{2^{\bm{y}}}{\bm{x}}-1, \frac{\bm{x}}{2^{\bm{y}}}-1\right)\right)
    \\
    \mathrm{AAcc}(\bm{y}, \bm{x}) &= \frac{\Big{|} \left|2^{\bm{y}} - \bm{x}\right| < 0.01\bm{x}\Big{|}}{|\bm{x}|} 
\end{align}

\section{Additional Probing Results}
\label{sec:additional-probing-results}

\begin{figure*}[ht]
    \centering
    \includegraphics[width=0.8\textwidth]{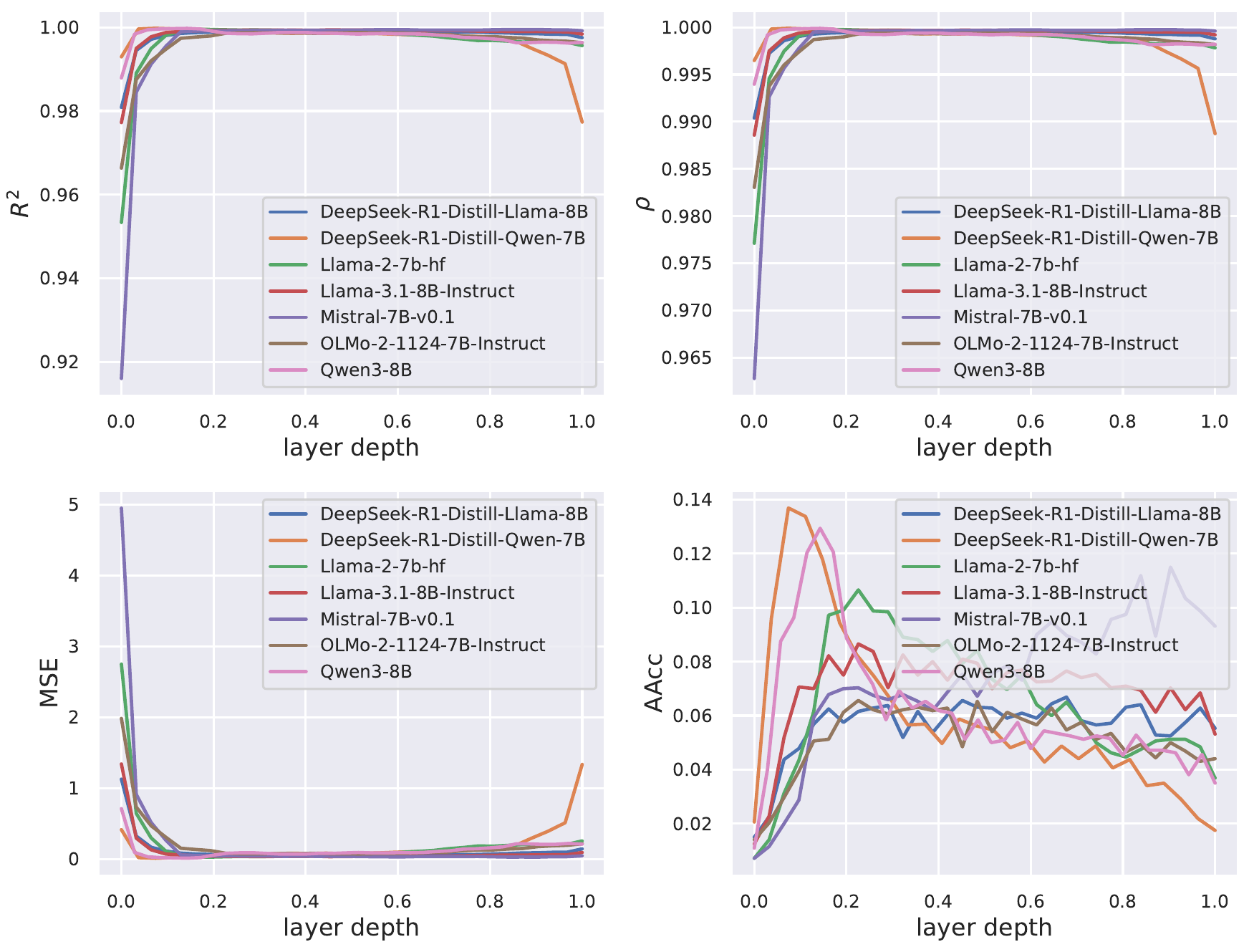}
   \caption{
   Regression probe performance on cross-notation data of each LLM across layers.  
   Here, we report the Pearson correlation $\rho$, the coefficient of determination $R^2$ (which $=\rho^2$ for linear regression), mean square error (MSE), and approximate accuracy (AAcc).  
   Definitions of MSE and AAcc can be found in \cref{appendix:metrics}.
   This is the expanded version of \cref{fig:regression_mse_layers}.  
   }
    \label{fig:regression_layers}
\end{figure*}

\begin{figure*}[ht]
    \centering
    \includegraphics[width=0.8\textwidth]{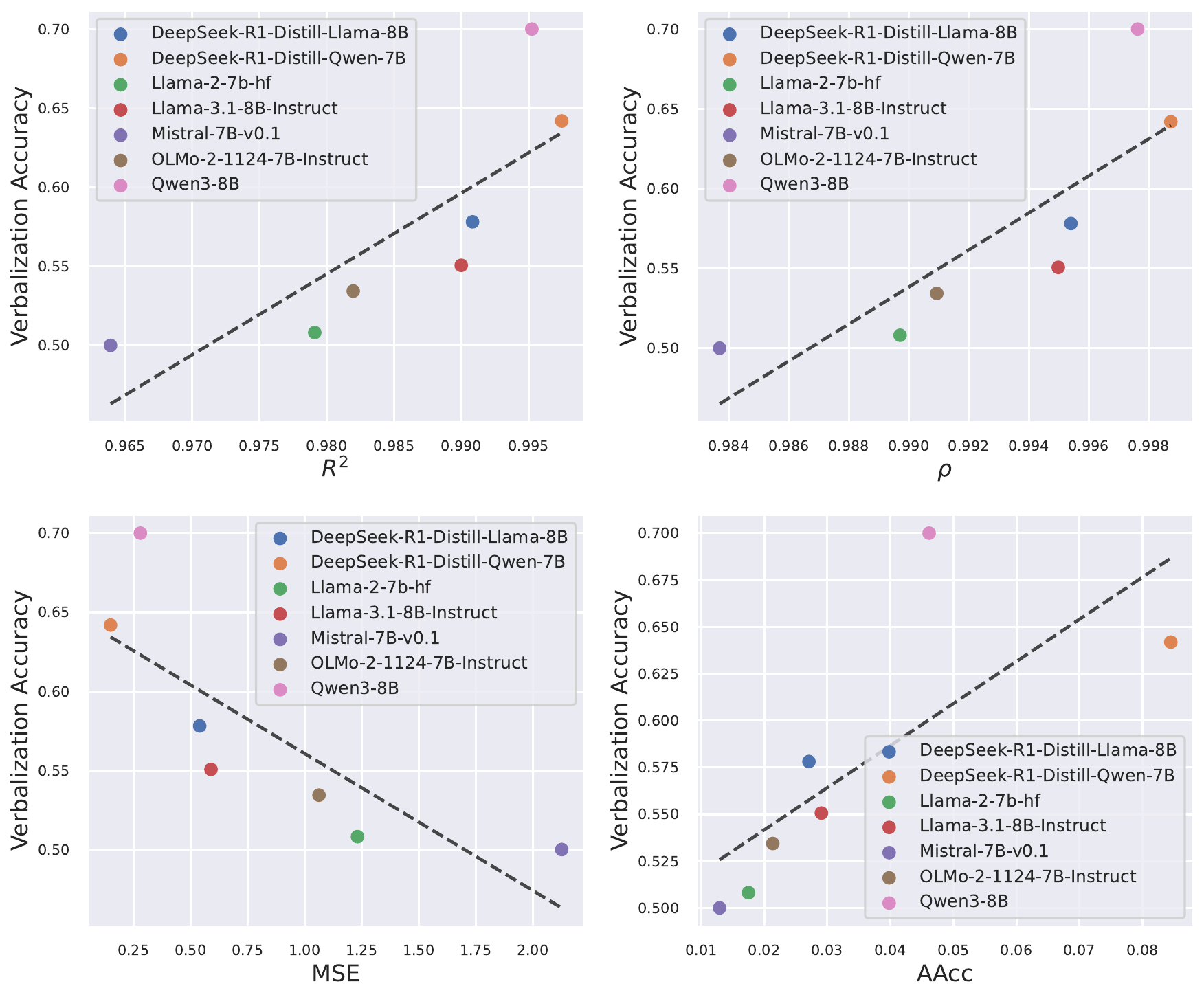}
    \caption{Average performance of linear regression probes at the first 3 layers correlates with model's verbalization accuracy on \texttt{cross-notation}.  This is an expanded version of \cref{fig:correlation-regression-mse}.}
    \label{fig:correlation-regression-all}
\end{figure*}

For a more detailed regression evaluation, we consider the Pearson correlation $\rho$, the coefficient of determination $R^2$ (which $=\rho^2$ for linear regression), mean square error (MSE), and approximate accuracy (AAcc).%
\footnote{See 
\cref{appendix:metrics} for more details about metrics.} 
\cref{fig:regression_layers}
and \cref{fig:correlation-regression-all} are expanded versions of 
\cref{fig:regression_mse_layers} and
\cref{fig:correlation-regression-mse}.

\begin{figure*}
    \centering
    \includegraphics[width=0.98\linewidth]{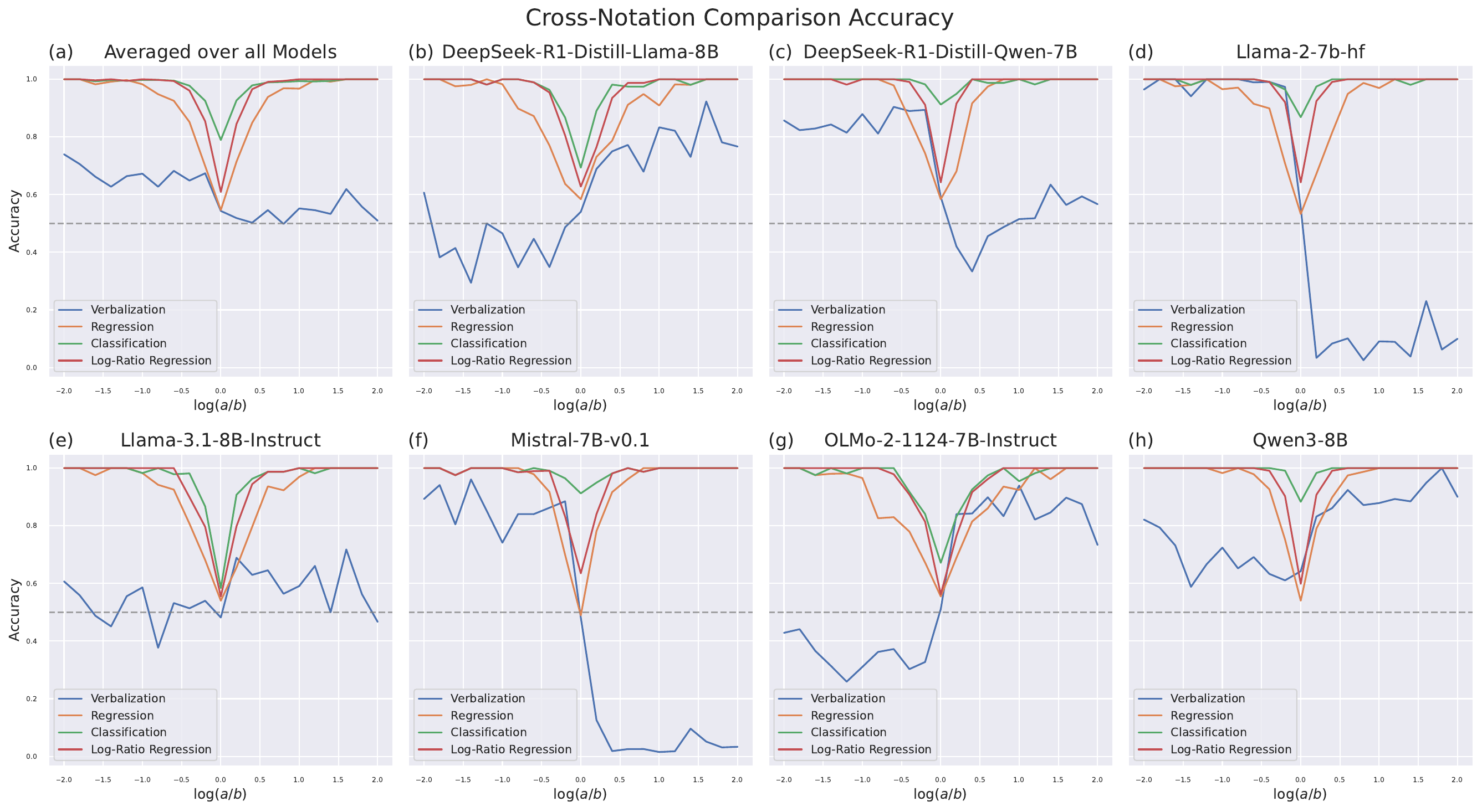}
    \caption{Cross-notation comparison accuracy with varying relative magnitude of two numbers (measured by $\log_2(a/b)$) using different methods.  Same as \cref{fig:log-ratio-plot}, but with each models' results displayed separately.}
    \label{fig:log-ratio-plot-individual-models}
\end{figure*}

The individual models' results for \cref{fig:log-ratio-plot} can be found in \cref{fig:log-ratio-plot-individual-models}.

\begin{figure*}[ht]
    \centering
    \includegraphics[width=0.8\textwidth]{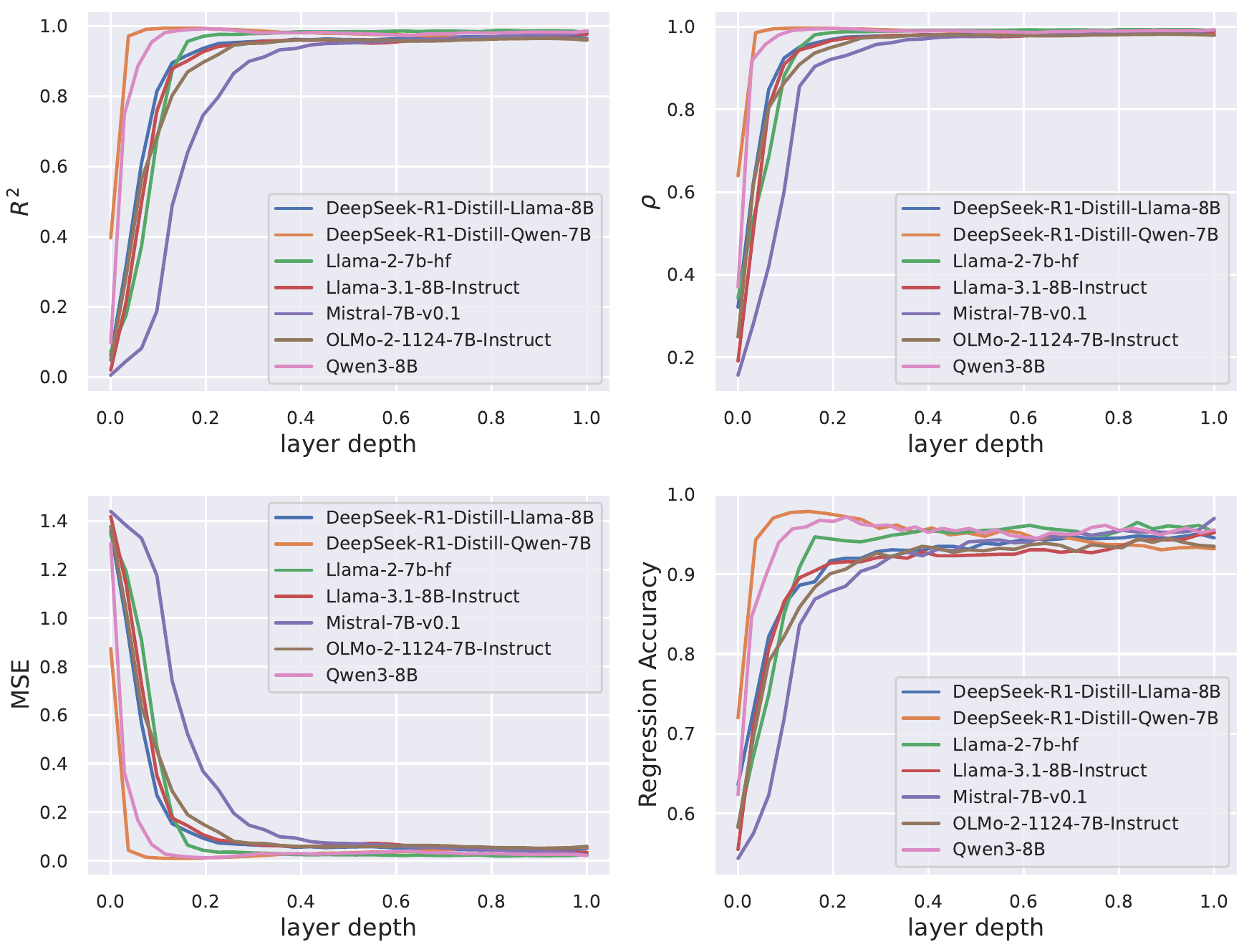}
    \caption{
    \textit{Log-ratio} regression probe performance on cross-notation data of each LLM across layers.  
    Here, we report the Pearson correlation $\rho$, the coefficient of determination $R^2$ (which $=\rho^2$ for linear regression), mean square error (MSE), and approximate accuracy (AAcc).  
    Definitions of MSE and AAcc can be found in \cref{appendix:metrics}.
    }
    \label{fig:log-ratio-correlation-regression-all}

\end{figure*}

\section{Additional Error Analysis}
\label{sec:additional-error-analysis}

Here, we present additional experimental results and analysis to complement our discussion in \cref{sec:verbalization-results}.\jason{Here are questions I asked earlier.  Are things any better with 10-shot?  How about finetuning?  I'm guessing not, but you should check.  Second, are the verbalized comparison results at a given level of $|\log a - \log b|$ actually any worse than what you'd get by comparing the probe outputs?  Or are they basically what you'd expect given the noisiness of the probe?  I might think they would be somewhat \emph{better} because the verbalized comparison can also try other strategies such as comparing digit-by-digit.}

\begin{figure*}
    \centering
    \includegraphics[width=0.98\linewidth]{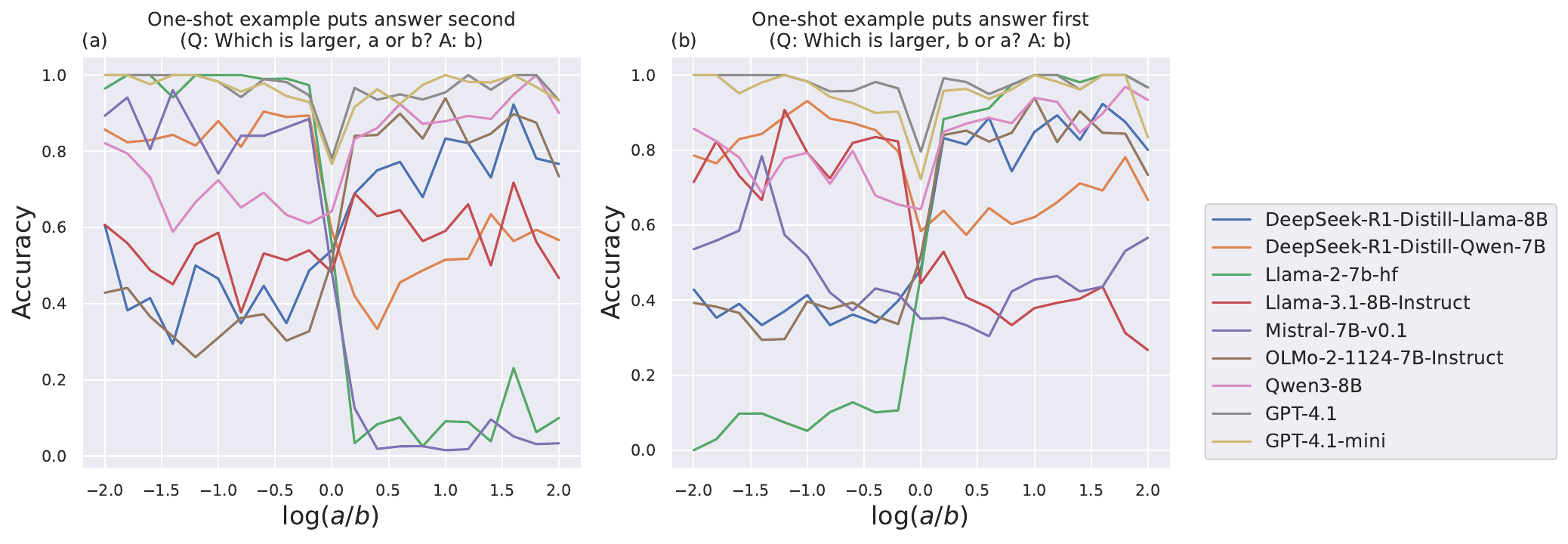}
    \caption{
    One-shot verbalized cross-notation comparison accuracy using different examples.
    (a) Same as \cref{fig:log-ratio-plot}(a); 
    (b) Models' verbalized response when the numbers in the example are exchanged in positions.
    }
    \label{fig:alt-prompt}
\end{figure*}

\paragraph{One-shot Prompting.}  
We observed in \cref{sec:verbalization-results} and \cref{fig:log-ratio-plot} that Llama-2-7B and Mistral-7B ignore the values of the numbers and answer only with the second number.  
This may be because the example that we provided in the one-shot prompt (see \cref{sec:experimental-setup,appendix:inference-setup}) placed the correct answer in the second position.
Thus, we tried exchanging the two numbers in the example.  \Cref{fig:alt-prompt} displays the results of the altered prompt next to the original.

We observe that, out of all models, Llama-2's response to the change in prompt is the most extreme.  
In both prompts, Llama-2 answered exclusively with the same position in which the answer was presented in the one-shot example.
Other models (DeepSeek-R1-Distill-Llama-8B, Mistral) show similar tendencies, but to a lesser extent.
OLMo-2 shows a slight though consistent preference for the second number, regardless of the one-shot prompt.
In contrast, GPT-4.1 and GPT-4.1-mini show minimal sensitivity to the ordering of the two options in the prompt, maintaining consistent performance across both variations.

\paragraph{Few-shot Prompting.}
Throughout this work, we have focused on one-shot prompting.  We
take the view that numeracy should be an innate capability: models
should already exercise basic numeracy skills---cross-notation
comparison being just one example---whenever they read scientific
documents.  In lieu of instructions, a single demonstration should be
enough to illustrate the intended task and output format.
Nevertheless, we evaluate few-shot performance here for completeness.

\begin{figure*}
    \centering
    \includegraphics[width=\linewidth]{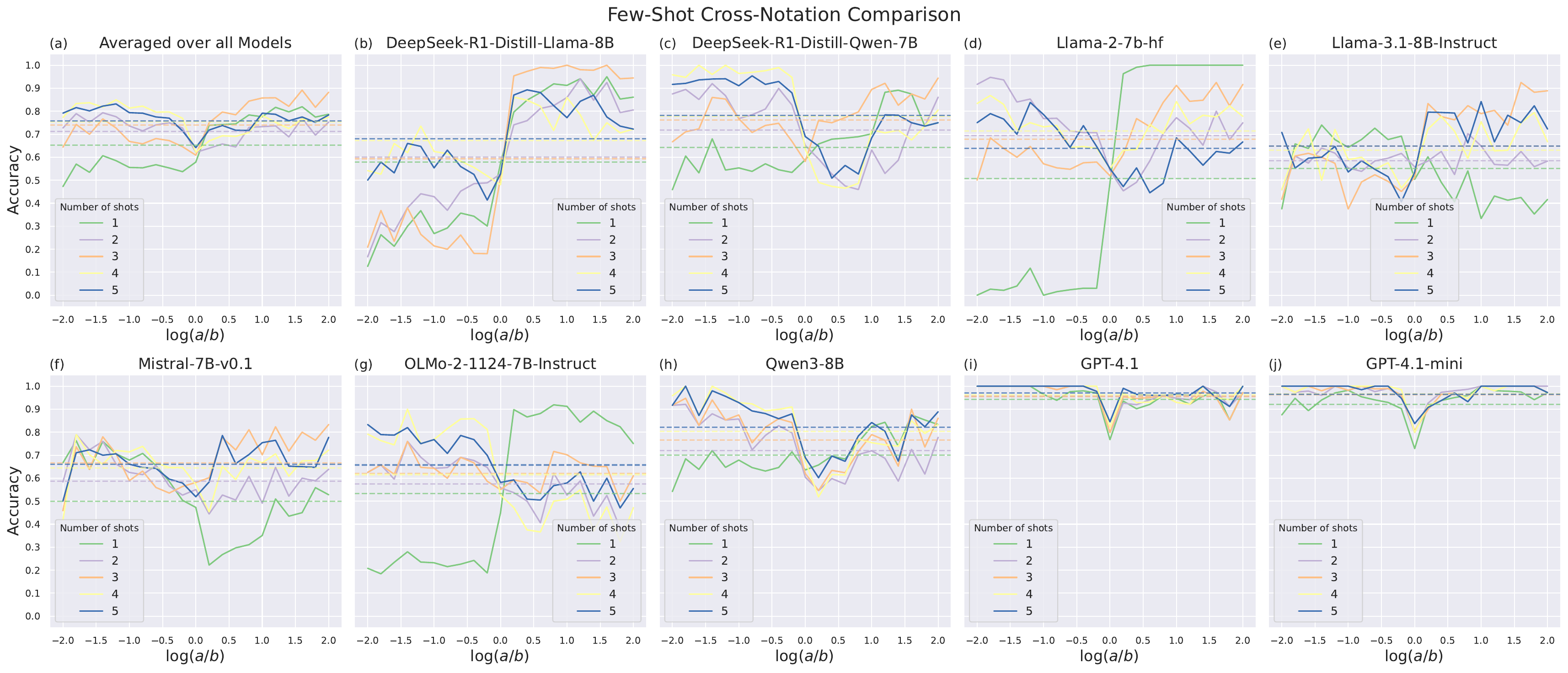}
    \caption{
    Few-shot verbalized cross-notation comparison accuracy for different models.
    In each panel, color-matched dashed lines indicate the average accuracy of each shot count.
    }
    \label{fig:few-shot-log-ratio-plot}
\end{figure*}

\cref{fig:few-shot-log-ratio-plot} shows the few-shot performance on cross-notation comparison using up to 5 examples (see specific few-shot prompts in \cref{appendix:inference-setup}).
Overall, all models benefit from additional demonstrations in the few-shot prompt, with performance improvement plateauing around 5 examples.
Still, all models struggle when the two numbers are close ($|\log_2(a/b)|<0.1$). While the smaller models (7B--8B) perform around random chance in this regime, GPT-4.1 and GPT-4.1-mini maintain 75--85\% accuracy, with additional demonstrations providing marginal further improvement.

We designed our 5 examples so that the correct answer's position alternates between the first place and the second place, in an attempt to eliminate positional bias we observed in \cref{fig:alt-prompt}.
While Llama-2, OLMo-2 and Mistral's positional biases are remedied by additional examples, the positional bias of Deepseek-R1-Distill-Llama-8B remains.

\begin{figure*}
    \centering
    \includegraphics[width=0.95\linewidth]{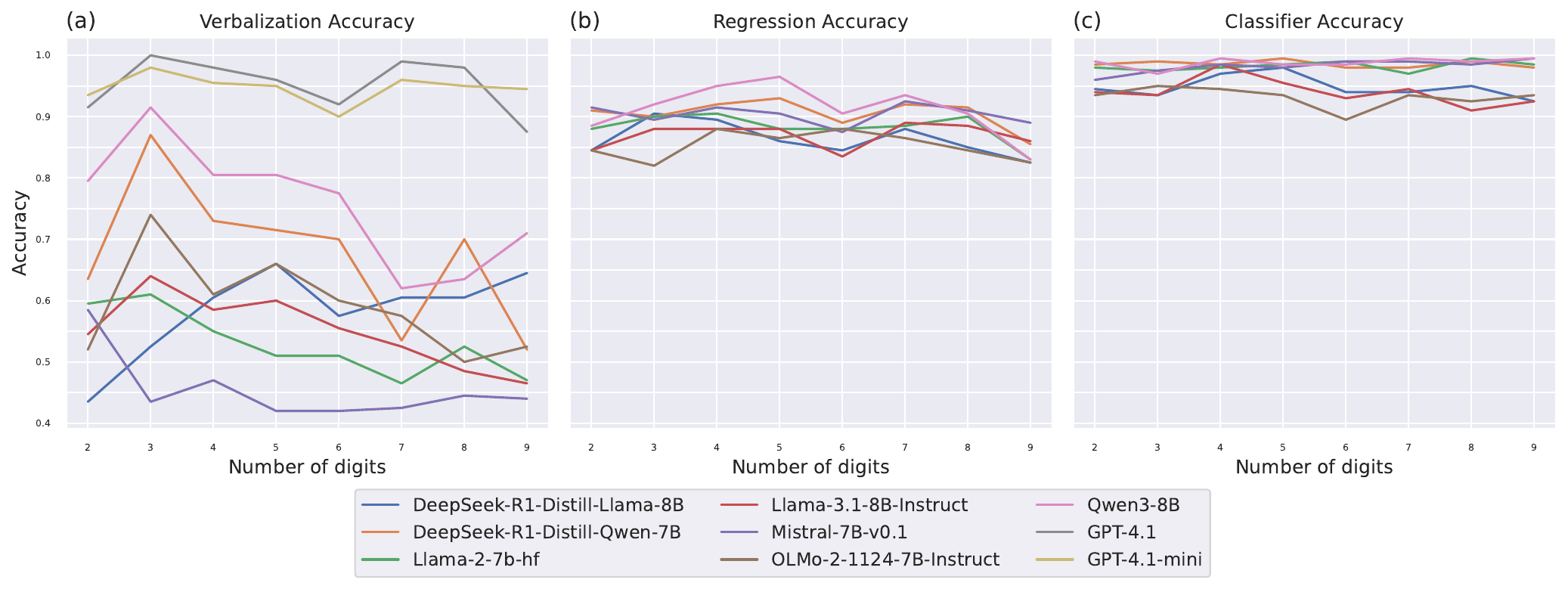}
    \caption{
    Cross-notation comparison accuracy vs.\@ number of digits using different comparison methods.
    (a) Verbalized comparison using one-shot prompting;  (b) Comparison via the predicted value of the regression probe trained on the hidden states; (c) Comparison using a logistic classification probe trained on the hidden states;
    }
    \label{fig:digit-plot}
\end{figure*}

\begin{figure*}
    \centering
    \includegraphics[width=0.95\linewidth]{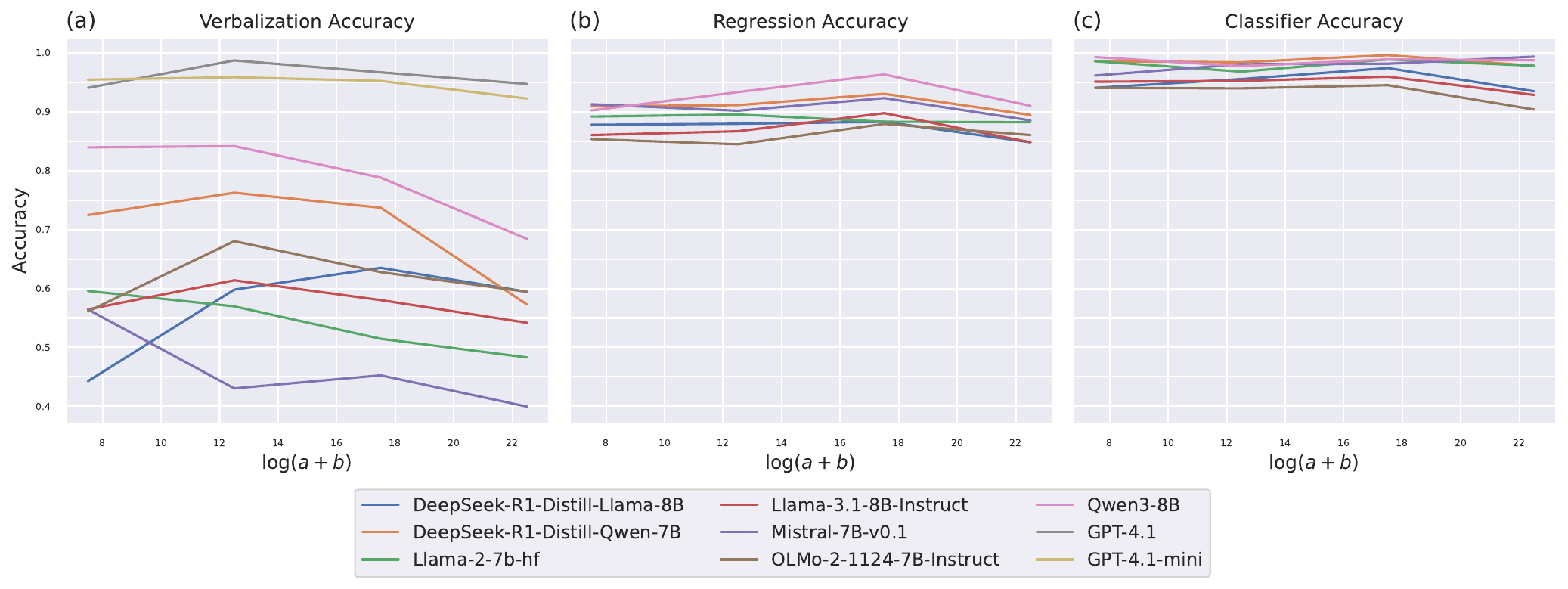}
    \caption{
    Cross-notation comparison accuracy vs.\@ $\log_2(a+b)$ using different comparison methods.
    (a) Verbalized comparison using one-shot prompting;  (b) Comparison via the predicted value of the regression probe trained on the hidden states; (c) Comparison using a logistic classification probe trained on the hidden states.
    }
    \label{fig:magnitude-plot}
\end{figure*}

\paragraph{Number of digits and magnitudes of numbers.}  Finally, we examine the effect of the number of digits and magnitudes of numbers on cross-notation comparison via either verbalization or probes.
\cref{fig:digit-plot} and \cref{fig:magnitude-plot} plot accuracy with respect to number of digits and magnitude (measured by $\log_2(a+b)$), respectively.
We observe that verbalized comparison accuracy generally deteriorates for numbers that are larger or have more digits in their surface form.
In contrast, comparison accuracy via regression or logistic classifier remains much higher and does not deteriorate in either case.
These observations suggest that LLMs implicitly understand larger and longer numerals but fail to verbalize their relationship.

\section{Computational Budget}\label{appendix:cluster}
 All our experiments used a single NVIDIA A100 GPU (40GB or 80GB).
Extracting hidden states for the full dataset takes a few hours, training and evaluating probes for one model takes about 40 minutes, and verbalization evaluation on 1,600 examples takes roughly 25 minutes per model. Finetuning a single model under one hyperparameter setting takes around 1 hour.  

\end{document}